# Deep Machine Learning Approach to Develop a New Asphalt Pavement Condition Index

**Hamed Majidifard[a], Yaw Adu-Gyamfi[b], William G. Buttlar[c]**

[a] Ph.D. Candidate, University of Missouri, Columbia, MO-65211

[b] Assistant Professor, University of Missouri, Columbia, MO-65211

[c] Professor and Glen Barton Chair in Flexible Pavements, University of Missouri, Columbia, MO-65211

**ABSTRACT:**

Pavement condition assessment provides information to make more cost-effective and consistent decisions regarding management of pavement network. Generally, pavement distress inspections are performed using sophisticated data collection vehicles and/or foot-on-ground surveys. In either approach, the process of distress detection is human-dependent, expensive, inefficient, and/or unsafe. Automated pavement distress detection via road images is still a challenging issue among pavement researchers and computer-vision community. In recent years, advancement in deep learning has enabled researchers to develop robust tools for analyzing pavement images at unprecedented accuracies. Nevertheless, deep learning models necessitate a big ground truth dataset, which is often not readily accessible for pavement field. In this study, we reviewed our previous study, which a labeled pavement dataset was presented as the first step towards a more robust, easy-to-deploy pavement condition assessment system. In total, 7237 google street-view images were extracted, manually annotated for classification (nine categories of distress classes). Afterward, YOLO (you look only once) deep learning framework was implemented to train the model using the labeled dataset. In the current study, a U-net based model is developed to quantify the severity of the distresses, and finally, a hybrid model is developed by integrating the YOLO and U-net model to classify the distresses and quantify their severity simultaneously. Various pavement condition indices are developed by implementing various machine learning algorithms using the YOLO deep learning framework for distress classification and U-net for segmentation and distress densification. The output of the distress classification and segmentation models are used to develop a comprehensive pavement condition tool which rates each pavement image according to the type and severity of distress extracted. As a result, we are able to avoid over-dependence on human judgement throughout the pavement condition evaluation process. The outcome of this study could be conveniently employed to evaluate the pavement conditions during its service life and help to make valid decisions for rehabilitation or reconstruction of the roads at the right time.

**KEYWORDS:** Pavement monitoring; Pavement distresses detection; Deep learning; Google API; Machine learning; Pavement condition prediction; YOLO; Image processing

# 1. Introduction

Pavement distress detection is a first key step in developing a robust pavement management system. It offers a comprehensive assessment of pavement conditions. Consequently, it generates information needed to make more cost-effective and consistent decisions associated with the pavement network preservation. Generally, pavement distress inspection is performed using sophisticated data collection vehicles and foot-on-ground surveys. In either approach, the current process of distress detection is human-dependent, expensive, inefficient, and unsafe. For example, the total price of an Aran was reported by the Ohio Department of Transportation for US$1,179,000, with an annual operating expense of US$70,000 [1]. Fully automated distress detection systems requiring no specialized data collection equipment, have the potential to lower distress survey costs, reliability and scalability [2]. The primary goal of this study is to leverage recent advances in machine learning to develop a low-cost, robust pavement condition assessment system, capable of detecting, classifying and quantifying the density of pavement cracks in an automated fashion.

The most promising approaches for automated distress analysis, leverage image processing and computer vision algorithms to detect edges of different types and severities of crack in pavement images. The primary advances in automated pavement crack detection techniques are as follow: intensity-thresholding [3-6], match filtering [7], edge detection [8], seed-based approach [9], multiscale methods like wavelet transforms and empirical mode decomposition [10-14]; texture-analysis, and machine learning [15-18]. Also, Zou et al developed CrackTree as an automatic procedure for crack detection [19]. There are some challenges related to these approaches. The first challenge is that these techniques rely on image pixel manipulations, which are very slow processes, and then it is not applicable for large scale deployment. Secondly, these techniques only work precisely if the image configuration is static, and the models don't work well if different camera configurations are used. Finally, there are many heuristic rules related to the use of these models, which make it impractical to be implemented extensively.

In order to overcome these challenges, computer vision algorithms which use machine learning models have been proposed as an alternative to traditional image manipulation techniques. In fact, recent progresses in deep learning has directed to substantial improvements in our ability to analyze streams of videos and image at unprecedented accuracies. The models are leading advances in areas like self-driving cars and smart cities [20], nanomaterials [21], healthcare, agriculture, retailing, and finance.

Deployment of machine learning approaches for automated pavement distress detection is not novel anymore, however application of deep learning is still attractive for pavement researchers [22-25]. Deep architecture with many hidden layers like deep convolution neural networks (DCNNs) provide frequent abstraction levels [26-29]. CrackNet software was established by Zhang et al as a crack detection model using raw image patches via the CNN [22]. Afterward, Zhang et al applied the Recurrent Neural Network (RNN) to produce CrackNet-R, which is more accurate in detecting small cracks and removing noise than CrackNet [30]. It must be noted that none of those mentioned above studies, proposed models based on a comprehensive dataset covering all pavement distress types from sections with different conditions. Also, classification and quantifying the density of the distress did not take into account simultaneously. Furthermore, none of these studies did not provide a pavement condition tool which can be used for evaluating pavement condition based on the proposed detection models.

In the current study, we develop four prediction models to evaluate pavement condition using the proposed pavement distress detection deep learning based models. It must be noted that the variety, quality and quantity of data utilized for training models is the key factor for their robustness, which is not always available. Comprehending the importance of labeled datasets to develop a strong pavement condition detection model, Majidifard et al represented the 'Pavement Image Dataset,' or (PID) [31]. The dataset contains 7,237 images extracted from 22 different pavement sections, including both US highways and interstate routes. A python code was developed to extract images using Google Application Programming Interface (API) in Google street-view [31]. Images were hand annotated by pavement experts carefully considering nine different distress classes. Afterward, the performance of dataset was assessed using a famous deep learning framework named You Look Only Once (YOLO v2) [31]. However, the proposed model by YOLO did not quantify the density of the cracks. Therefore, in our new study we developed a new U-Net based model to quantify the density of the distresses. Finally, various pavement condition indices tried to be developed based on the proposed crack detection models (Yolo-based and U-Net based). The following summarizes the primary contributions of this study (Figure 1):

- First, we introduced a unique dataset annotated for simultaneous classification and densification of pavement distresses. The data is extracted from google street view, which provides us with a variety of camera views needed to improve the system's ability to recognize different types of cracks and estimate their severity.
- Second, we implemented a distress segmentation model capable of delineating the boundaries of different types of cracks in challenging environments characterized by severe shadowing, non-crack-like objects such as cars, trees, etc. We overcome this challenge through intelligent model training techniques and architectural designs.
- Developing a new pavement condition index based on inputs from a crack classification model (YOLO), a density model (U-Net) and a hybrid of machine learning-based models.
- Conduct a comparative analysis of proposed pavement condition index with PASER ratings along pavement profile.

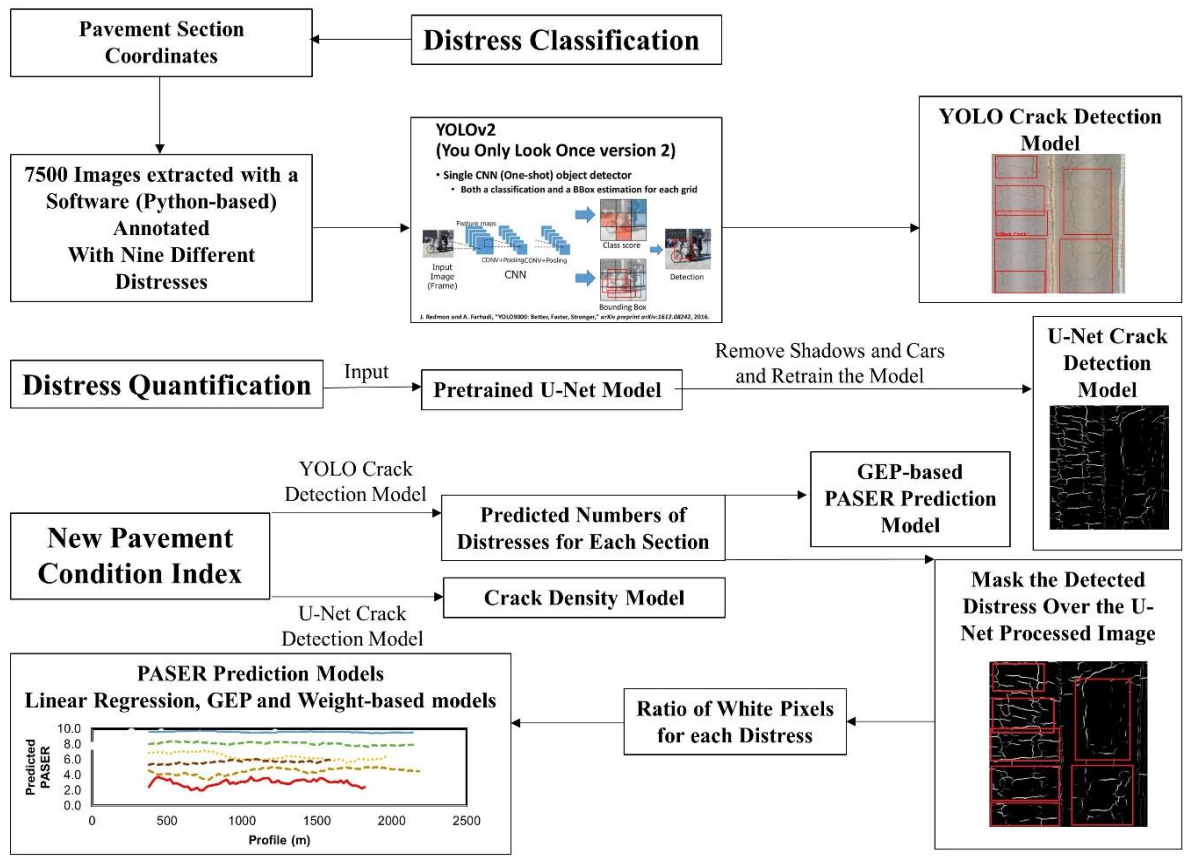

*Figure 1. Experimental research flowchart*

The arrangement of the article is organized as follows. The relate works including pavement condition indices and automated crack detection methods are reviewed in Section 2. The methodology of the research introduces main road segmentation, U-Net based Model for distress quantification, and developing pavement condition indices which discussed in section 3 followed by model validation briefed in Section 4. Section 5 center around the conclusion of the research which summaries the methodology and the implementation of the study.

## 2. Background

### 2.1. Pavement condition index development approaches

There are various indices that are utilized to characterize pavement conditions, such as the pavement condition index (PCI), present serviceability index (PSI), present serviceability rating (PSR), pavement surface evaluation and rating (PASER) [32]. PCI is the quantitative-based pavement condition index, while the others can be grouped into class of qualitative pavement indices. The PCI, arguably the most prevalent index for pavement condition evaluation, depends on inspection data and visual observation. U.S. Army Corps of Engineers developed PCI for the management of pavement rehabilitation and maintenance system [33]. In the PCI calculation method, various distress types with different severities are incorporated into a univalent PCI value. The length or area and the severity (low, medium, and high) of each distress are taken into account to calculate the deduct values for the surveyed section (34; 35). The PCI ranges from 100 to 0, in which 100 is brand new pavement, and 0 is the worst condition possible. The University of Wisconsin-Madison Transportation Information Center developed the PASER system, rates pavement conditions from 1-10 . It utilizes visual inspection to assess pavement surface conditions [36]. The type of distress is not required to calculate PASER, as a result, PASER ratings cannot be disaggregated by distress types. The advantage of this method is that pavement sections may be rated promptly, possibly even by "windshield survey." A principal drawback is that the PASER outputs cannot be used in mechanistic-empirical transportation asset management programs because PASER ratings cannot be classified into component distress data [36].

Newer approaches for developing pavement condition indices have been proposed by several research teams as a result of the limitations traditional methods such as the PCI, PASER, PSI, etc. Eldin and Senouci used neural-network based algorithm to predict condition of pavement using data provided by the Oregon State Department of Transportation (ODOT) [37]. Fathi et al used a hybrid machine learning (ML) method that combines random forest (RF) and artificial neural network (ANN) to predict the alligator deterioration index (ADI) [38]. Piryonesi and El-Diraby used historical distress data in the LTPP database to develop decition trees-based algorithm to predict PCI of asphalt roads [39]. They developed a Python™ program to produce the PCI values from distress data according to the ASTM methodology. For this purpose, all deduct value graphs, and correction curves were digitized and mathematically designated [39]. Shahnazari et al., used ANN and GP algorithms to predict the PCI. The models established based on large dataset of PCI values which was collected from road network of Iran. The proposed models were accurate and they can be used instead of Micropaver software to calculate PCI [40].

The limitation of the previous models is that all of them rely on manual inspection of distresses. Inspectors should calculate the area, length, and severity of distresses, and the output can be used as an input in these models to calculate the pavement condition rate. Therefore, a fully automated distress detection model is demanding.

### 2.2. Automated Crack detection, segmentation and classification approaches

**Image segmentation** is the procedure of subdividing an image into several segments (sets of pixels, also known as image objects). Image segmentation is classically performed to detect boundaries such as lines and curves in images. In another word, image segmentation is the practice of extracting the interest objects from the background. **Object classification** is a technology associated with image processing and computer vision that copes with detecting semantic objects of a specific class such as humans, buildings, or cars in videos and images. Object classification and segmentation – both are part of machine learning based image processing to train AI algorithms through computer vision, and both are important for object recognition precisely in machine learning and AI development. The classification process is easier than segmentation; in the classification approaches, all objects in a single image are grouped or categorized into a single class. While in segmentation each object of a single class in an image is highlighted with different shades to make them recognizable to computer vision. In crack segmentation, the severity of the distresses is detectable although there is no ability to classify them into different groups. On the contrary, in the classification approaches the distresses can be categorized into different groups and the severity cannot be measured.

The primary segmentation approaches are thresholding and edge detection. Thresholding-based segmentation is broadly utilized in automated pavement distress systems [41]. Another regular procedure in image processing is edge detection. The significant advantage of edge detection is the fast reduction of image data to beneficial information. There are many useful edge detectors that have been proposed over the past 30 years, such as LOG, Sobel, Roberts, and Prewitt edge detectors [42]. The main problem associated with the most edge detection algorithms is that these algorithms only characterize a spatial scale for edges detection. Pavement images are acknowledged to be challenging to work within the process of pavement distresses because of different details at various scales. In the past decade, wavelet-based edge detection at multiple scales became popular in pavement image processing [43]. Shadows and lighting effects in pavement images introduced new challenges in the automatic pavement distress detection field. Region-based image thresholding has been implemented to resolve the difficulties caused by shadow and illumination variations. A neighboring difference histogram procedure was used by Li and Liu to crack image segmentation using a globally optimized threshold [44]. However, histogram-based procedures, do not consider photometric and geometric characteristics of the cracks in road pavement images [45]. Dynamic local thresholding for non-overlapping image blocks, developed by Oliveira and Correia [4]. Although segmentation procedure based on thresholding is beneficial in various image segmentation tasks, it is still problematic for the automatic threshold selection. Image morphological procedure is another primary tool that has often been utilized in the automated pavement distress detection studies [46]. Naoki and Kenji [47] developed a procedure using fundamental top-hat transform and morphological operations to extract structural information from road pavement image, and subsequently cracks detection. Although morphological image processing provides the benefit of extracting prominent geometrical structure related to the cracks in road surface images, the performance is highly parameter dependent [3]. In practice, it is suggested to operate morphological processing along with other image processing techniques.

**Object classification** generally divided to machine learning-based and deep learning-based methods. In Machine Learning approaches, first the features defined by using Histogram of oriented gradients (HOG), Viola–Jones object detection framework, Scale-invariant feature transform (SIFT) [48], then classification frameworks such as support vector machine (SVM) implemented to do the classification. While, deep learning techniques have the capability to do end-to-end object detection without precisely defining features, and are classically based on convolutional neural networks (CNN) such as You Only Look Once (YOLO) [49], Single-Shot Refinement Neural Network for Object Detection (RefineDet) [50], Region Proposals (R-CNN [51], Fast R-CNN [52], Faster R-CNN [53], Single Shot MultiBox Detector (SSD) [54]. Critical distresses on road surface must be detected in order to propose a strong automated pavement distress detection model. Pavement deterioration rate is a function of various features such as climate, structural layering, traffic, and layer age. Strategies are developed by road administrators to repair the road surface based on the type, extent, and severity of the distresses. Former researches made improvement in the direction of this target but falling short in one area or another. To clarify, CrackNet [22; 55] focused on detecting the presence of cracks but did not specifically identify individual types. Zalama et al. [56] classified the distress types horizontally and vertically, while Karaköse et al. [57] categorized distresses into three classes – vertical, horizontal, and alligator. Finally, other studies resulted in the blurry road markings detection [58], while others concentrated on cracks classification, including cracks which were sealed [59]. Quality of data used for training machine learning models is the essential key for their robustness. Previous studies introduced several benchmarked datasets (private and public) for training of machine learning models [60]. In spite of this, none of the aforementioned studies employed an inclusive dataset covering all distress types which are annotated cautiously by pavement experts.

Until recently, very few studies have been performed to address the issue of fine-grained classification of pavement distress. The accuracy of distress classification is dependent on factors such as camera view angle, segmentation accuracy, etc. The adoption of recent advances in machine learning has dramatically improved the robustness of distress classification methods. However, despite the great achievement of deep learning models, existence of shadows and poor lighting and low contrast have made it challenging for pavement researchers to develop an intact automated pavement distress detection model. Most of current models only work accurate when the cracks are discernible, generally with salient gray-level and uniform illumination features. Also, the camera angle view parameter is a challenging issue in crack segmentation and classification. Most of the trained crack classification models only work precisely if the same camera angle view images are used for prediction.

Furthermore, the former studies did not try to develop a model to classify and quantify distress density simultaneously. In section 2.3, we review our previous research, which we introduced a dataset with both top-down and wide-view images which are available in GitHub repository [61].

## *2.3. Developing Crack Detection Model using Deep-Learning Framework (Majidifard et al., 2020)*

The current study is inspired by Majidifard et al's recent work which developed a comprehensive dataset for training deep learning algorithms for classifying different types of pavement distress. This section highlights their work and its relevance to the current study. The study first defined and annotated nine types of the most critical distresses which affect pavement condition selected after reviewing various studies [62-70]. Next, a large database of pavement images attained from 22 various pavement sections in the United States via a python-based software which communicate with API of Google street view. All the distresses on the acquired images were then annotated using the pre-defined distress classification rules as shown in Table 1. Figure 2 offers instances of images annotated with nine different distress types. As shown in the table 1, reflective, lane longitudinal, sealed longitudinal, and block cracks are among the highest number of boundary boxes and images found in our dataset. Two different images were extracted at each coordinate location: Wide-View images with a pitch angle of -70° and birds-eye-view at -90°. The wide view images were found to be beneficial for classification of distresses, whereas the top-down view image resulted in more precise distress severity quantification.

*Table 1. Distress types versus their corresponding distress ID [31].*

| Distress Type | Distress ID | Image Example | Number of boundary boxes for each class |
|---|---|---|---|
| Reflective Crack | D0 | 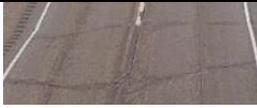 | 12428 |
| Transvers Crack | D1 | 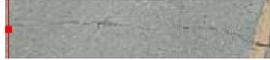 | 5343 |
| Block Crack | D2 | 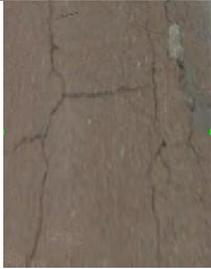 | 9709 |
| Longitudinal Crack | D3 | 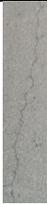 | 2892 |
| Alligator Crack | D4 | 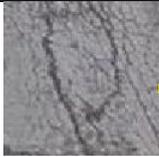 | 4838 |
| Sealed Reflective Crack | D5 | 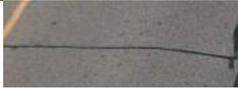 | 5910 |
| Lane Longitudinal Crack | D6 | 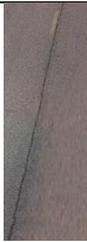 | 12102 |
| Sealed Longitudinal Crack | D7 | 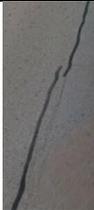 | 13610 |
| pothole | D8 | 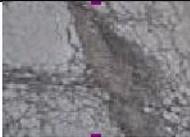 | 637 |

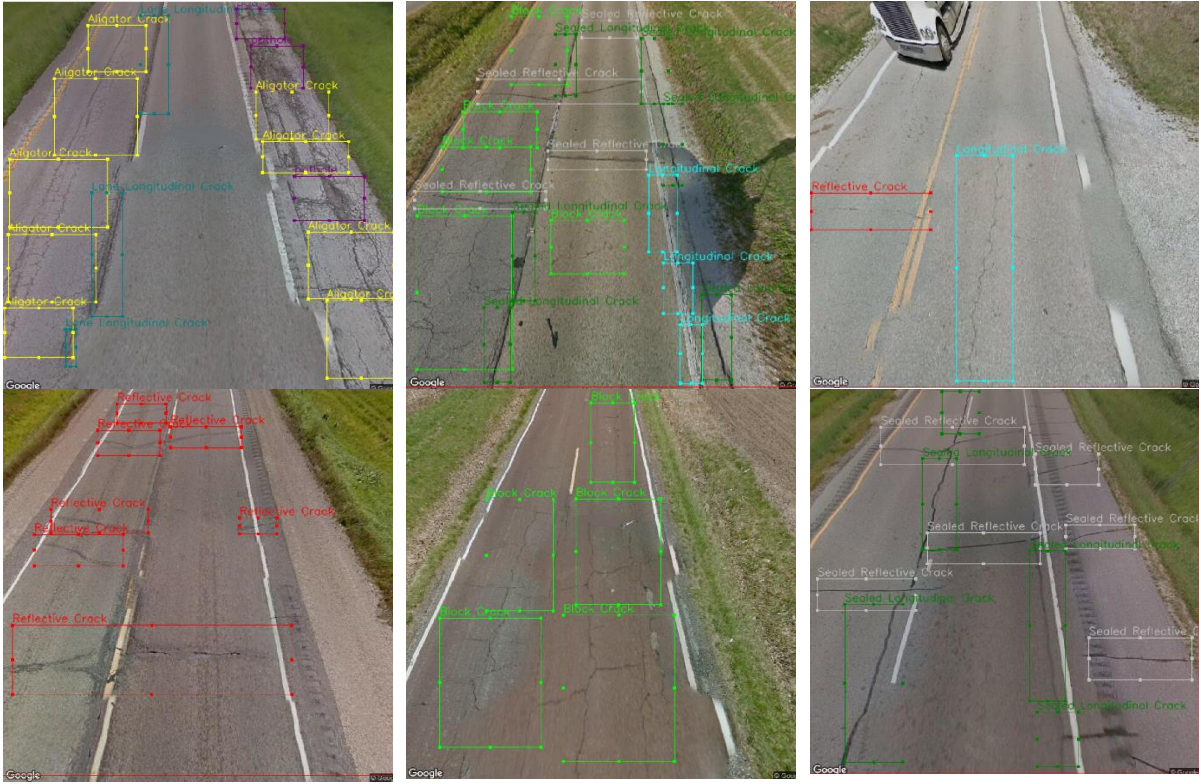

*Figure 2. Annotated images (wide-view images) in the PID dataset [31].*

An advanced deep learning framework named YOLO v2 was employed to detect and classify nine pavement distress types, automatically.

Figure 3 shows the architecture of the YOLO model that was applied. The reader is directed to the original paper for model development and testing details (Figure 3).

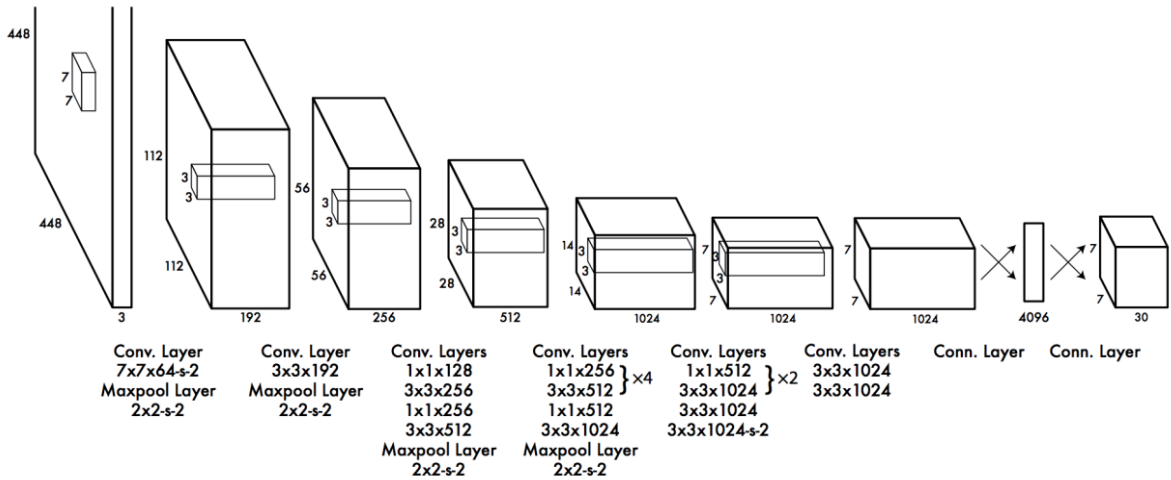

*Figure 3. YOLO architecture [49]*

**Figure 4** shows some examples of prediction results on top-down images. To challenge the robustness of the model, full sunshine images and images containing shadows (trees and cars) were selected. Although the model was trained on wide-view images, the prediction results on top-down images, confirm that the developed model has the capability to detect distresses in both the shadow-containing and full sunshine images, accurately.

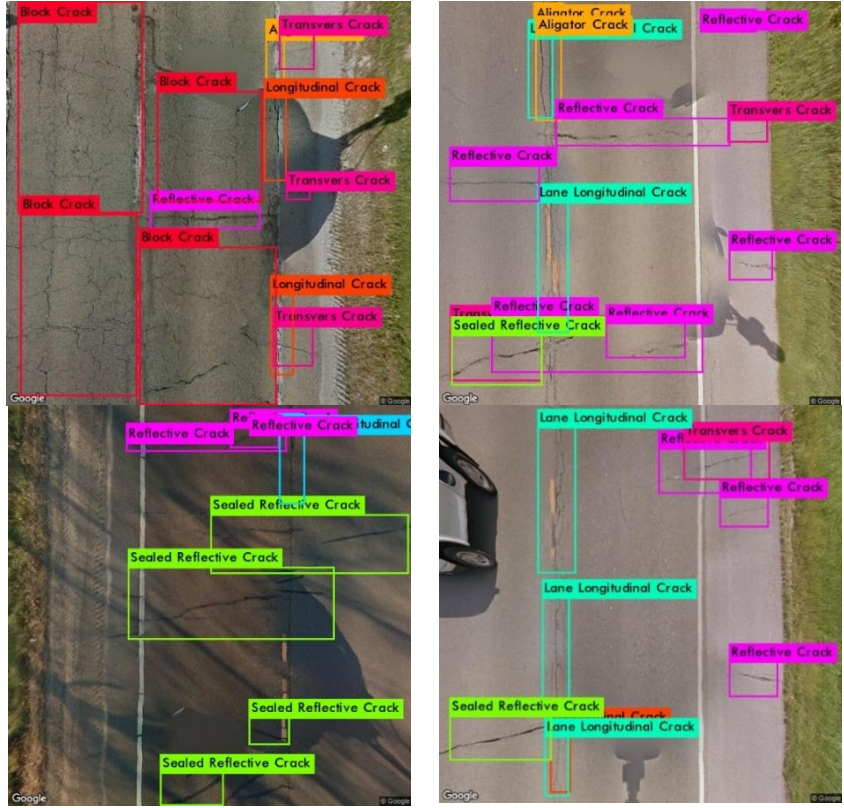

*Figure 4. Samples of detecting pavement distresses from top-down images [31].*

Precision, Recall, F1 scores and confusion matrices were the parameters utilized to assess accuracy of the model.

Table **2** represents detection and classification accuracies of the proposed model for the nine classes in our dataset. The precision, recall and F1 values for the model were 0.93, 0.77 and 0.84, respectively [31].

**Table 2.** Detection and classification results for nine distress types [31].

| Crack class name | D0 | D1 | D2 | D3 | D4 | D5 | D6 | D7 | D8 | **Average** |
|---|---|---|---|---|---|---|---|---|---|---|
| **Precision** | 0.93 | 0.9 | 0.93 | 0.91 | 0.91 | 0.93 | 0.93 | 0.94 | 0.96 | **0.93** |
| **Recall** | 0.76 | 0.83 | 0.79 | 0.83 | 0.74 | 0.83 | 0.79 | 0.57 | 0.78 | **0.77** |
| **F1** | 0.84 | 0.86 | 0.85 | 0.87 | 0.82 | 0.87 | 0.85 | 0.71 | 0.86 | **0.84** |

In this section, we reviewed approaches to develop pavement condition models and crack detections methods. Also, we explained why we introduced a new pavement image dataset, which was used to develop a deep-learning-based distress detection tool. In the following sections we highlight the main contributions of this study: first, we introduce a robust pavement distress segmentation and densification approach based on deep convolutional neural networks. Second, we develop a pavement condition index based on inputs from distress class types and density information. To the authors' knowledge, none of the previous studies have been able to rate the condition of pavements by taking into account the type and extent of distresses surveyed. Detecting and quantifying of distresses are not enough to evaluate the pavement conditions. As mentioned in the literature review section, PASER is one the pavement condition index which helps to rate roads consistently. However, PASER index is a qualitative and human dependent index, we noticed that there are some mis-ranking and inconsistency between ranking of the sections. **Figure 5** shows example of inconsistency in PASER rating among the investigated sections.

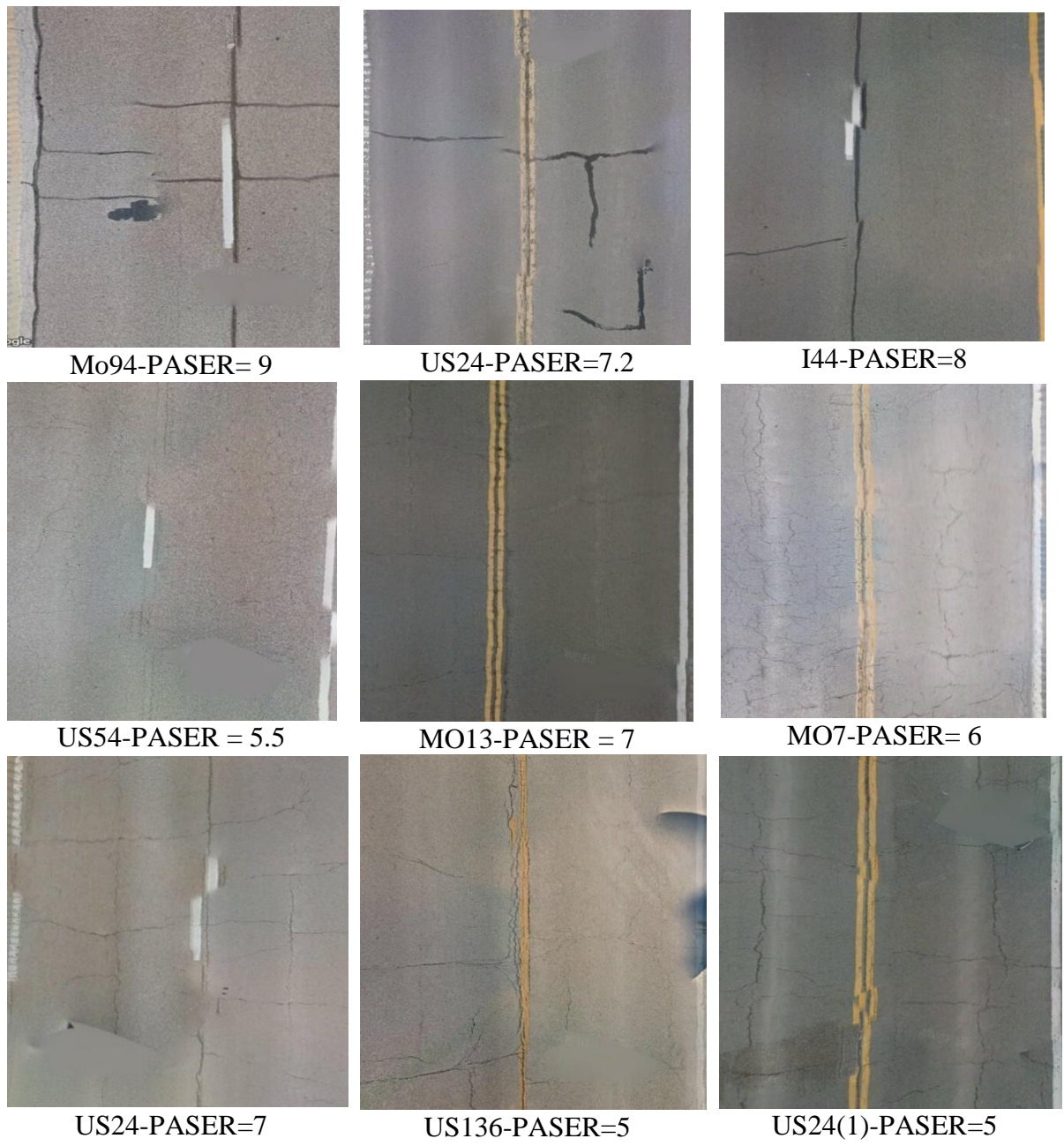

*Figure 5. Example of inconsistency in PASER rating among the investigated sections*

In order to develop a robust prediction tool, we need to have a reliable response. Then, the PASER values were revised by pavement experts and various approaches were implemented to fit a model with the eight features to the response value.

**3. Methodology of Developing Pavement Condition Prediction Models**

The methodology of developing the pavement condition prediction models consists of four main steps: First, 71 pavement sections were selected in the state of Missouri, and the PASER values were extracted from MoDOT virtual portal. The PASER values were checked and corrected in case of a discrepancy by pavement experts. Afterward, an average of 83 images per section was extracted from the Google map at corresponding GPS coordinates using the developed python software. Second, shoulders were removed from images by road segmentation technique, and the revised images were run through the developed YOLO model to classify the distresses. Third, a U-Net based model was developed for distress quantification. Forth the YOLO and U-net models merged together to create the hybrid model, and finally, three prediction PASER models were developed using YOLO and hybrid models by implementing various machine learning techniques. Figure 10 shows the flowchart of the three prediction models.

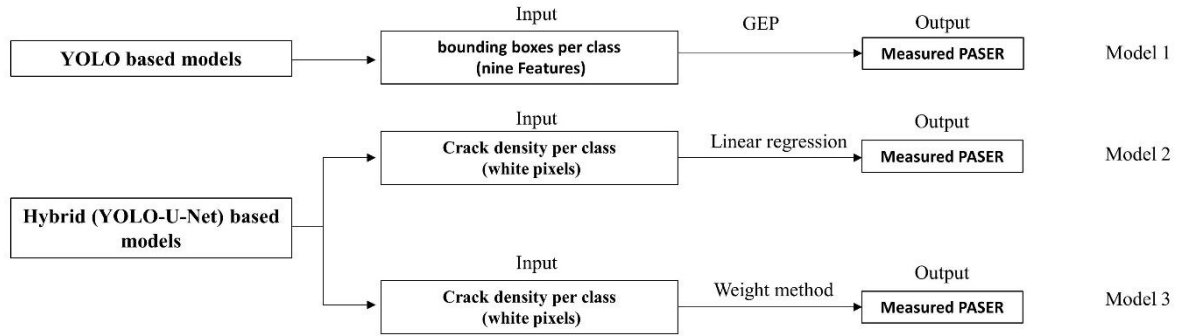

**Figure 6.** The primary PASER prediction models developed in this study.

### 3.1. Main Road Segmentation

The types of mixtures used for road shoulder construction usually vary significantly from those used for the main road. As a result, the types of distress and the underlying factors influencing them are also different. In the current study, distress information on road shoulders are not used in calculating the pavement condition index. We annotated and developed a shallow network for cropping out the main road for each pavement image. Distress detection and segmentation and subsequently carried out on the cropped image. Figure 5 shows examples of main road cropping.

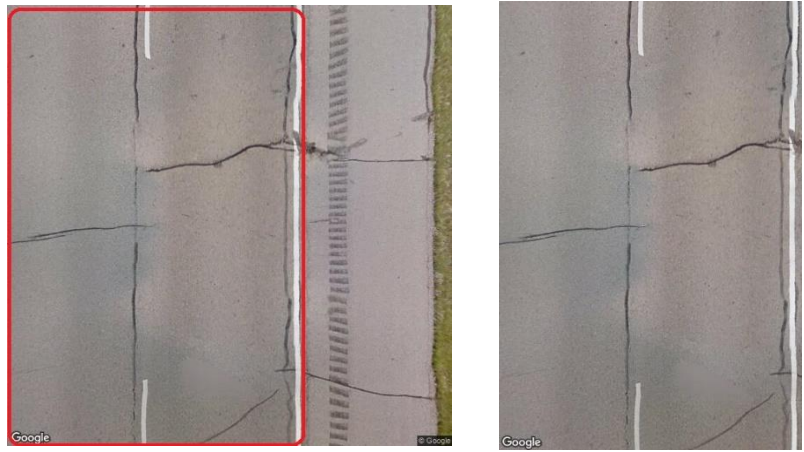

*Figure 7. Automatic cropping of the main-road from shoulder via CNN*

### 3.2. Developing a U-Net based Model for Distress Quantification

In the current study, a pre-trained U-Net convolutional network was used, which was originally developed for biomedical image segmentation [71]. Herein, it was used to quantify the density of cracks in roads.

Khanal and Estrada implemented a Neural baseline that does image segmentation to retinal vessel images. The dataset included 40 retinal images (20 for training and 20 for testing) where blood vessels were annotated at the pixel level to mark the presence (1) or absence (0) of a blood vessel at each pixel (i, j) of the image (Figure 9). The authors used the U-net architecture to do blood vessel segmentation (Figure 8). It is an architecture that is widely used for semantic segmentation tasks, especially in the medical domain. The U-net Architecture is an encoder-decoder with some skip connections between the encoder and the decoder. The ability to consider a wider context when making a prediction for a pixel is the primary benefit of this algorithm. This comes from the large number of channels utilized in the up-sampling process. Figure 8 shows the network architecture. The left and right side of the structure represents contracting and an expansive path, respectively. The contracting path reflects the typical architecture of a convolutional network. It contains the duplicated application of two 3×3 convolutions (unpadded convolutions), each accompanied by a rectified linear unit (ReLU) and a 2×2 max pooling operation with stride 2 for downsampling. At each downsampling step, the number of feature channels is doubled. The expansive path consists of an upsampling of the feature map followed by a 2×2 convolution that divides feature channels. This feature is connected with the correspondingly cropped feature map from the contracting path, and two 3×3 convolutions, each followed by a ReLU. At the final layer, a 1×1 convolution is utilized to map each 64×64 element feature vector to the desired class numbers. Overall, the network contains 23 convolutional layers [72].

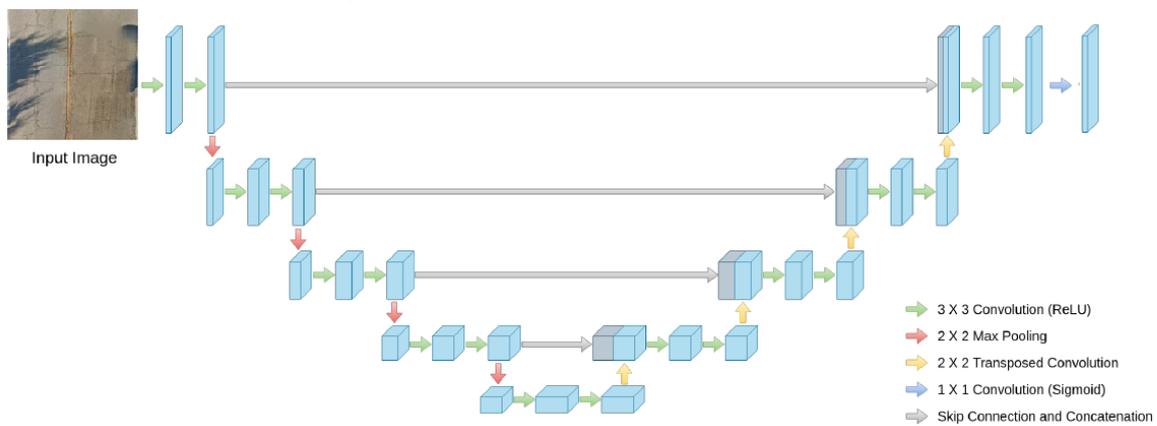

*Figure 8. U-net Architecture [72].*

After processing pavement images with the pretrained model we found it very accurate. However, the problem with this model is that it detects some external items like cars and shadows as distresses. In order to address the mentioned issue, shadows and cars were removed from the U_Net images with hand and the model retrained again. Figure 10 shows the images with shadows, cars and cracks. The images in the first, second and third columns represent original input, pretrained model and retrained model output, respectively. As seen in the Figure 10, cracks were detected accurately in the re-trained model although shadows covered them in some cases. Also, the shadows and cars were removed and the noise decreased significantly in the new model.

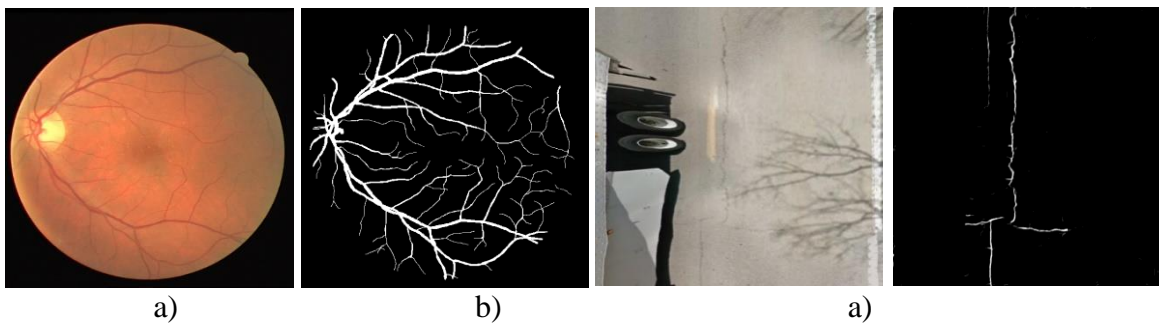

a)　　　　　　　　　　b)　　　　　　　　　　a)

*Figure 9 a. One example of the training data to developed the U-net model, b. hand-crafted annotations that were used to retrain the model.*

***Figure 10.*** *Examples of detecting shadows and cars as cracks a) raw image b) pre-trained model output, and c) retrained model output.*

In order to check the precision of the model, 20 images were selected for testing the model performance. The pixel difference between the ground truth and the predicted image were calculate as MSE (mean square error). **Figure 11** represents GT, predicted and error for one image in test dataset. The final MSE calculated as 0.25 for the 20 images in the test dataset.

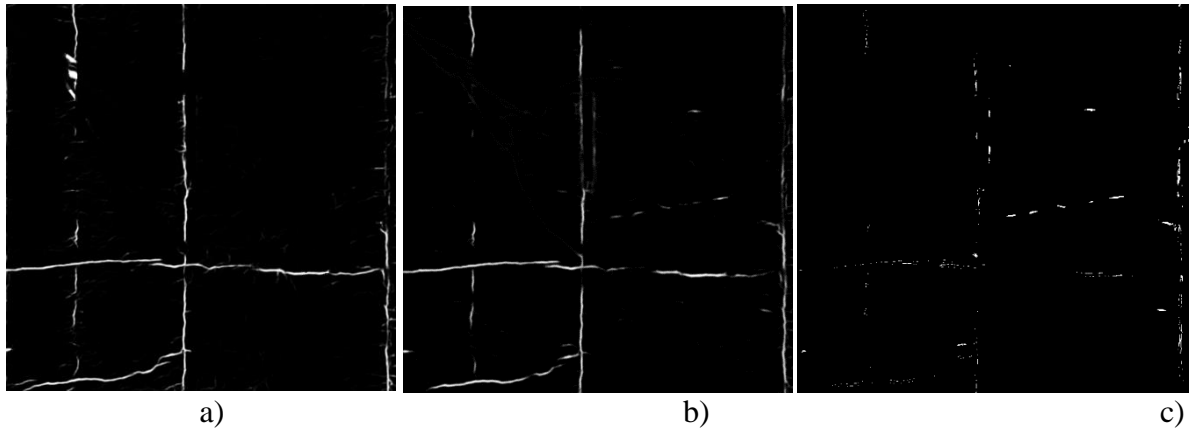

| a) | b) | c) |

*Figure 11. Image of a) ground truth, b) predicted and c) error.*

### 3.3. Developing a Hybrid Model of YOLO and U-Net

In order to develop a robust and comprehensive pavement condition model we needed to consider both type and density of the cracks. As mentioned before, none of the YOLO and U-net models could not provide the most accurate pavement condition model. Therefore, both of these models require to be integrated together. To address this concern, images were processed through the proposed YOLO and U-net models individually and the detected objects in the proposed yolo model image were masked on the corresponding images processed by the U-net model. Afterward the density of cracks were calculated for each type of distresses (**Figure 12**). The ratio of white pixels for each detected distress to the total image size are considered as the new features for our hybrid pavement condition model, which will be explained in detailed in section 3.4.3 and 3.4.4.

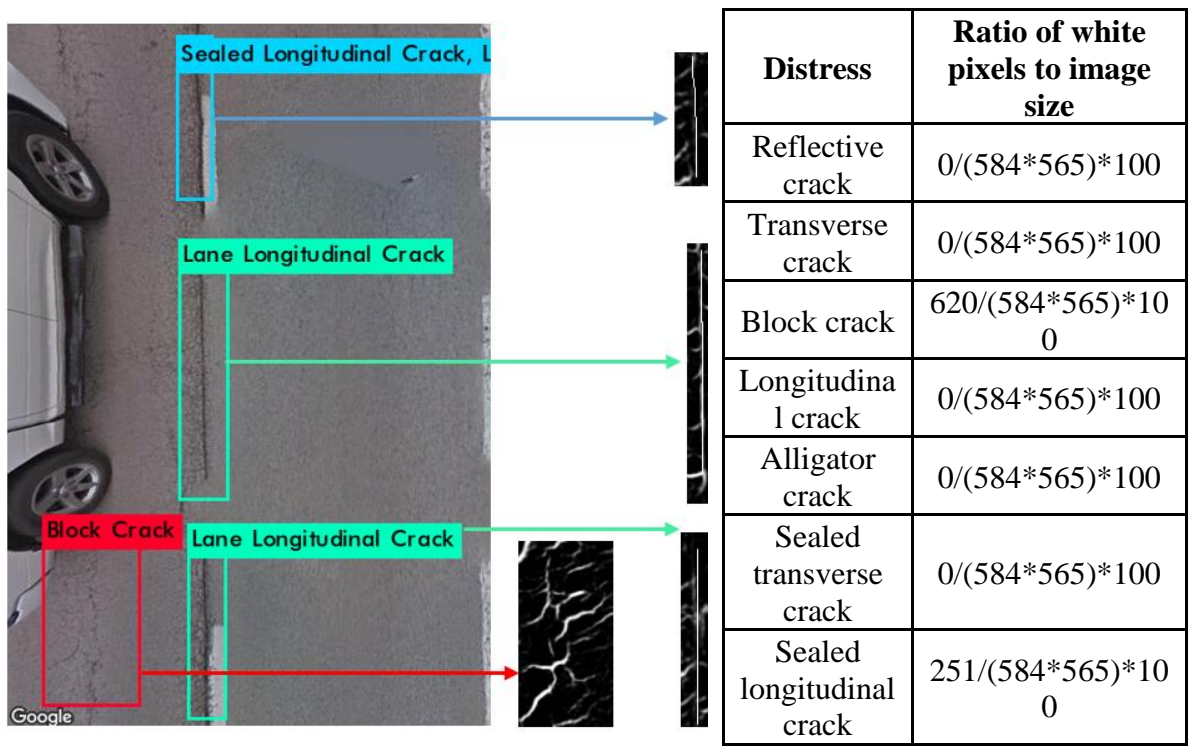

| Distress | Ratio of white pixels to image size |
|---|---|
| Reflective crack | 0/(584*565)*100 |
| Transverse crack | 0/(584*565)*100 |
| Block crack | 620/(584*565)*100 |
| Longitudinal crack | 0/(584*565)*100 |
| Alligator crack | 0/(584*565)*100 |
| Sealed transverse crack | 0/(584*565)*100 |
| Sealed longitudinal crack | 251/(584*565)*100 |

| Lane longitudinal crack | 521/(584*565)*100 |
|---|---|
| Pothole | 0/(584*565)*100 |

*Figure 12. Integrated Yolo and U-Net based model and calculation of ratio of white pixels*

### 3.4. Developing Pavement Condition Prediction Models

*3.4.1 Pavement Condition Prediction Model Development based on YOLO Model outputs Using Genetic Expression Programming*

In order to develop pavement condition prediction model, all the extracted images for each section were run through the developed YOLO crack detection model. The average numbers of detected distresses were calculated for each section. **Table 3** exhibits the detailed variables statistics used in this study. The data (71 pavement sections) are presented by frequency histograms (see **Figure 13**). As seen in the **Figure 13**, the distributions of the predictor variables are not uniform.

*Table 3. Statistical parameters of the dependent and independent variables.*

|  | D(0) | D(1) | D(2) | D(3) | D(4) | D(5) | D(6) | D(7) | D(8) | PASER |
|---|---|---|---|---|---|---|---|---|---|---|
| Mean | 0.58 | 0.26 | 0.82 | 0.21 | 0.11 | 0.20 | 0.41 | 0.33 | 0.00 | 7.2 |
| Median | 0.43 | 0.13 | 0.33 | 0.18 | 0.04 | 0.01 | 0.35 | 0.09 | 0.00 | 7.0 |
| Mode | 0.02 | 0.00 | 0.00 | 0.00 | 0.00 | 0.00 | 0.01 | 0.00 | 0.00 | 7.0 |
| Range | 2.09 | 1.07 | 4.32 | 0.94 | 0.92 | 4.07 | 1.37 | 2.65 | 0.08 | 6.0 |
| Maximum | 2.09 | 1.07 | 4.32 | 0.94 | 0.92 | 4.07 | 1.37 | 2.65 | 0.08 | 10.0 |
| Minimum | 0.00 | 0.00 | 0.00 | 0.00 | 0.00 | 0.00 | 0.00 | 0.00 | 0.00 | 4.0 |

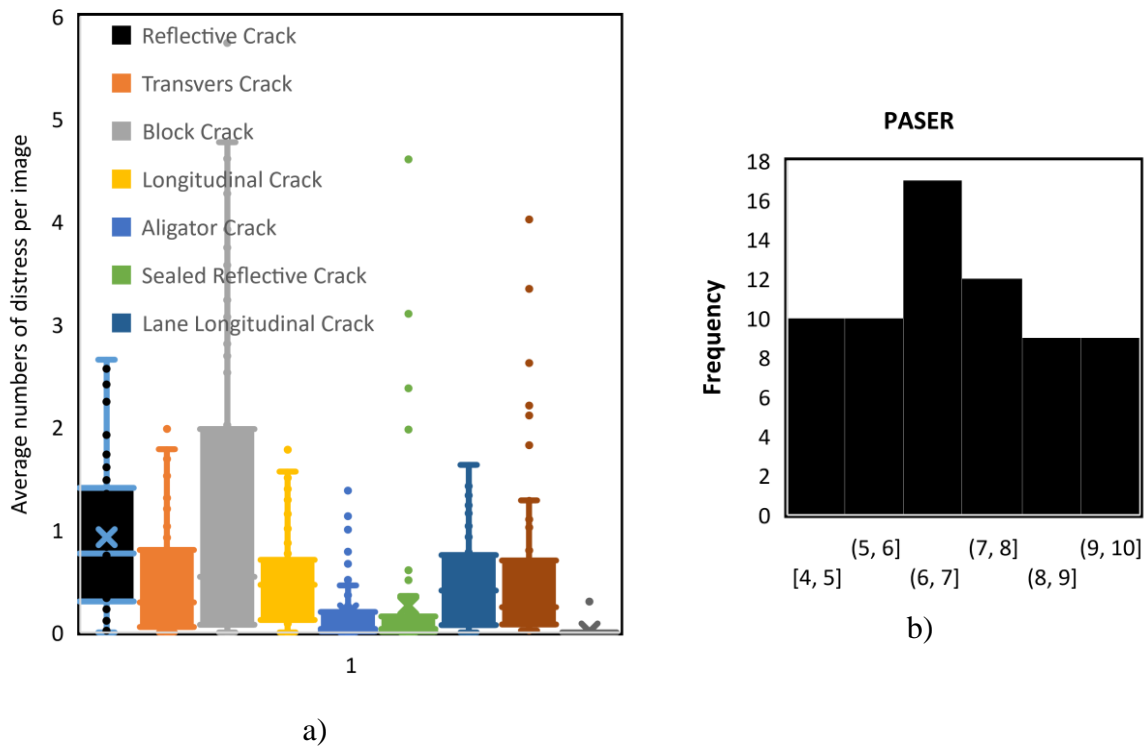

*Figure 13. Distribution histograms of a) the features and b) output.*

GEP is known as the most advanced generation of traditional genetic programming (GP) to generate nonlinear prediction models automatically. These techniques have been extensively deployed for tackling complicated engineering problems [73-80]. The traditional GP creates computational models by imitating the biological evolution of living organisms. It provides a tree-like form of a solution, which represents the closed-form solution of the optimization problem [81; 82]. The primary objective of GP is to generate a population of programs that correlate inputs with outputs for each data point. The initial random functions are then calibrated to seize fitting functions that can accurately estimate the output via an administered trial and error methodology.

The GP algorithm has some advantages over other machine learning (ML) techniques [83] such as adaptive neuro-fuzzy inference systems and artificial neural networks. The first benefit is that GP is not a black box and outputs are in the form of semi-complex mathematical solutions. The other advantage is that GP-based models lie on their inclination to obtain precise relationships without considering former patterns of the existing relationships. On the other hand, the primary advantage of GEP over the traditional GP method is that it can compile several sub-programs to create a single complex program. Moreover, the GEP algorithm can be implemented significantly faster than GP due to evolving the binary bit patterns. However, more fundamental details about the ML, GP and GEP methods can be found in [81; 82; 84]. In the current study, GEP is deployed for developing the rutting prediction models.

To avoid overfitting, the whole dataset (71 sections) was divided randomly into three categories as training (70%), validation (15%), and testing (15%). The training and testing subsets were used to calibrate and evaluate the models, respectively, while validation dataset was used as an external output to check the model performance.

This study presents a new ML-based model which predicts the PASER from the following variables:

$$PASER = f[d(1), d(2), d(3), d(4), d(5), d(6), d(7), d(8), d(9)] \quad [1]$$

where,

d(1)= Reflective Crack, d(2)= Transverse Crack, d(3)= Block Crack, d(4)= Longitudinal Crack, d(5)= Alligator Crack, d(6) = Sealed Reflective Crack, d(7)= Lane Longitudinal Crack, d(8)= Sealed Longitudinal Crack, d(9)= Pothole. All the variable including d(1) to d(9) are the average number of distresses per each section.

The model was developed using PASER values obtained from 71 pavement sections in the Midwest United States. Various runs were performed to delineate the optimized GEP parameters. There are various principal setting GEP parameters such as general setting, complexity increase, genetic operators, numerical constant, and fitness function. In general part, the number of chromosomes changes the simulation run time. The higher number of chromosomes, the longer running time. The head size represents the complexity of terms in the developed model. **Table 4** shows a set of parameters used during the GEP simulations.

*Table 4. The optimal parameter setting for the GEP algorithm.*

| Parameter | Settings | |
|---|---|---|
| General | Chromosomes | 30 |
| | Genes | 6 |
| | Head size | 12 |
| | Linking function | Addition |

| Functions | Function set | +, -, ×, /, √, ³√, Ln, Log, power, exp ,sin, cos, tan, |
|---|---|---|
| Complexity increase | Generations without change | 2000 |
| | Number of tries | 3 |
| | Max. complexity | 5 |
| Genetic operators | Mutation rate | 0.00138, 0.044 |
| | Inversion rate | 0.00546 |
| Numerical constants | Data type | Floating-point |
| | Lower bound | -10 |
| | Upper bound | +10 |

The optimal GEP-based prediction model for PASER is as follows:

$$Y1 = arctan((d(9)\wedge(1/3) - ((d(5) - d(3))\wedge 2)))); \qquad [2]$$

$$Y2 = sin(sin(tanh(((d(9) + (min((d(8) - d(1)), min(G2C4, d(6))) + (G2C7 + G2C7))) - d(4)))));$$

$$Y3 = (exp(min(d(6), (((d(3) + G3C8)\wedge(1/3) + d(2))/exp(d(6))) - (G3C6 - (1.0 - d(1))))));$$

$$Y4 = min(d(9), sin((min((exp(d(5)) * min(d(5), d(1))), (G4C2 - d(2))) + d(6))));$$

$$Y5 = (1.0/(((((G5C3 + G5C2) + (d(3) + d(1))))\wedge(1/3) * ((d(7) + G5C9))\wedge(1/3) + exp(reallog(max(d(6), G5C4))))));$$

$$Y6 = (1.0 - (min(d(2), G6C6) - (((1.0/((G6C9 - d(4)))) * ((d(3) + d(1)) + d(3))) - d(9))));$$

$$Y = Y1 + Y2 + Y3 + Y4 + Y5 + Y6$$

Where:

G2C7 = -0.19, G2C4 = 0.51, G3C6 = 9.60, G3C8 = 10.07, G4C2 = 5.40, G5C9 = 2.57, G5C4 = 1.83, G5C3 = 6.68, G5C2 = -7.18, G6C6 = -6.74, G6C9 = -3.36

d(1)= Reflective Crack, d(2)= Transverse Crack, d(3)= Block Crack, d(4)= Longitudinal Crack, d(5)= Alligator Crack, d(6) = Sealed Reflective Crack, d(7)= Lane Longitudinal Crack, d(8)= Sealed Longitudinal Crack, d(9)= Pothole.

**Figure 14** represents measured versus predicted PASER values for the entire data. Although the dataset size was small, an acceptable performance for the proposed model was achieved (**Figure 15**).

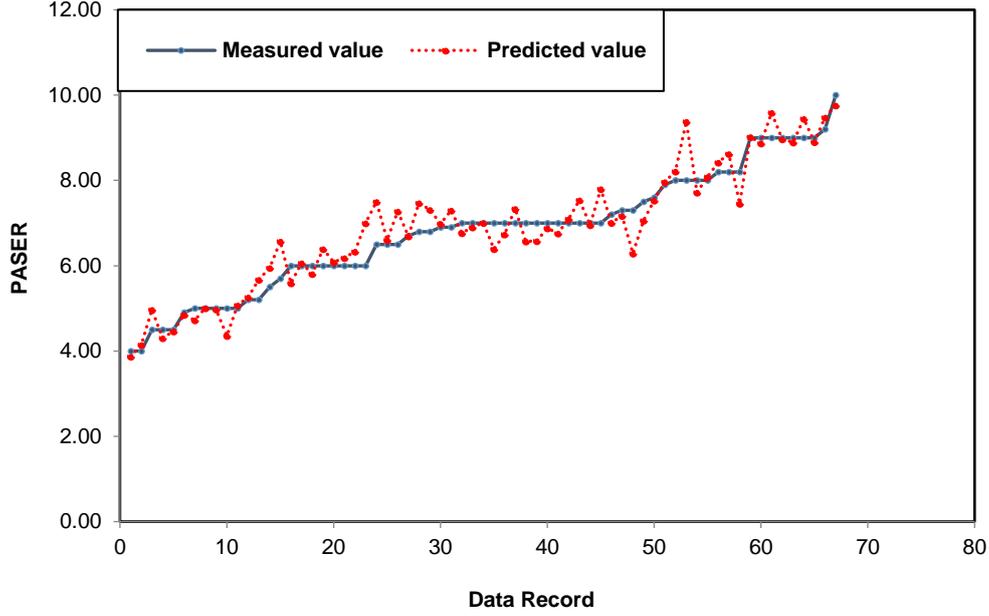

*Figure 14. Measured versus predicted value for the entire data.*

In order to measure the performance of model, coefficient of determination ($R^2$), root mean squared error (RMSE) and mean absolute error (MAE) were used.

$$R^2 = \frac{(\sum_{i=1}^{n}(O_i-\overline{O_i})(t_i-\overline{t_i}))^2}{\sum_{i=1}^{n}(O_i-\overline{O_i})^2 \sum_{i=1}^{n}(t_i-\overline{t_i})^2} \quad [3]$$

$$RMSE = \sqrt{\frac{\sum_{i=1}^{n}(O_i-t_i)^2}{n}} \quad [4]$$

$$MAE = \frac{\sum_{i=1}^{n}|O_i-t_i|}{n} \quad [5]$$

Where,
$O_i$: Measured value
$t_i$: Predicted value
$\overline{O_i}$: Average of measured values
$\overline{t_i}$: Average of predicted values
$n$: Samples number

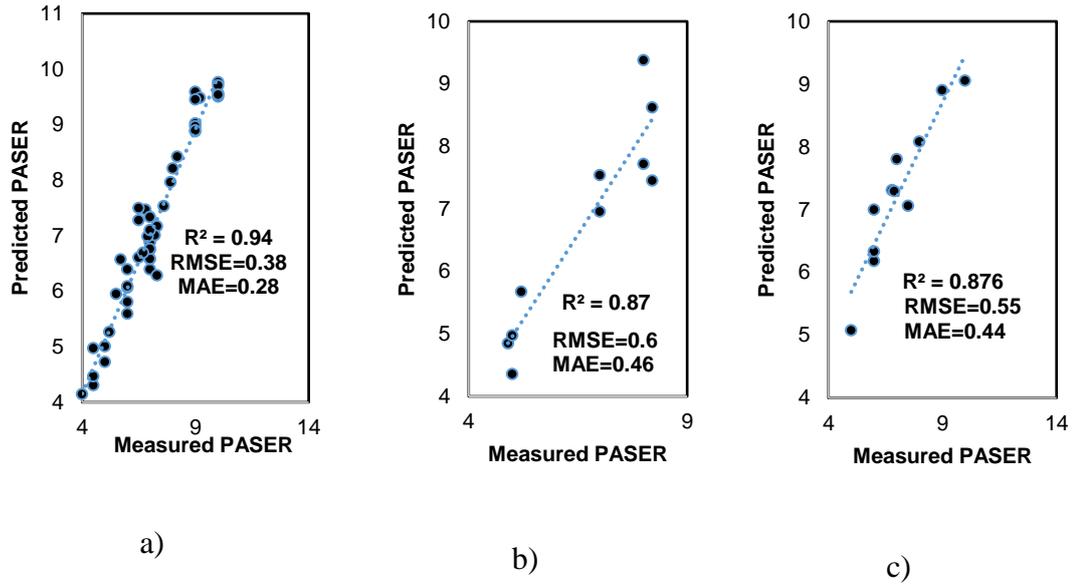

a)   b)   c)

*Figure 15. Measured against predicted PASER using the GEP model: (a) training data; (b) testing data, and; (c) validation data.*

A sensitivity analysis was carried out to explore the relative importance of variables in the GEP model. The results are presented in **Figure 16**. The major observations were:

- Block cracks had the highest influence on the pavement condition (PASER).
- Lane longitudinal, sealed longitudinal cracks and potholes had the lowest influence on PASER. The small effect of potholes on pavement condition may seem nonsense at first glance. However, none of the sections in our dataset have potholes on their surfaces. Therefore, the actual effect of potholes on pavement condition ignored in our prediction model. The same explanation is valid for the alligator and transverse cracking.

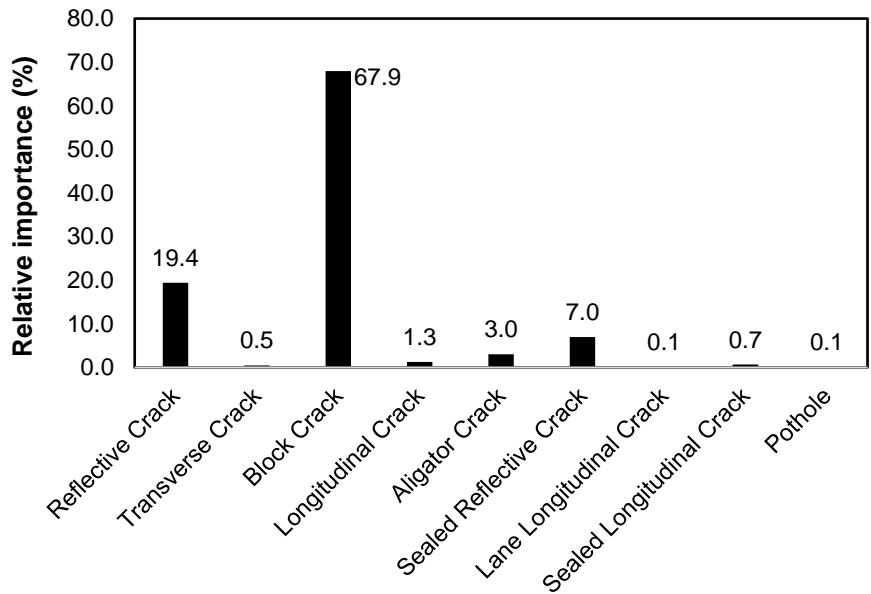

*Figure 16. Variable importance in the GEP model.*

*3.4.2. Pavement Condition Prediction Model Development based on Unet Model outputs*

In this part of study, a new pavement condition index was developed based solely on the U-net model output. All the 71 sections images were processed through the proposed U-net model. For each processed image, the ratio of white pixels (crack-like features) to the total size of the image was used to measure the crack density. To reduce the effect of spurious, non-crack-like features, the output of the U-Net mode was thresholded. Two thresholds were tested: the first threshold only selects pixels with intensity values ranging from 127 to 255, and the second one is from 200 to 255 counts as white pixels. The ratio of the white pixels were calculated for all sections and it is reported as the distress index to rank the sections.

**Figure 17** shows the correlation of the distress index from U-net model with corresponding PASER values. As shown in the figure, PASER has stronger correlation with the crack density threshold of 127 than the one with 200 threshold.

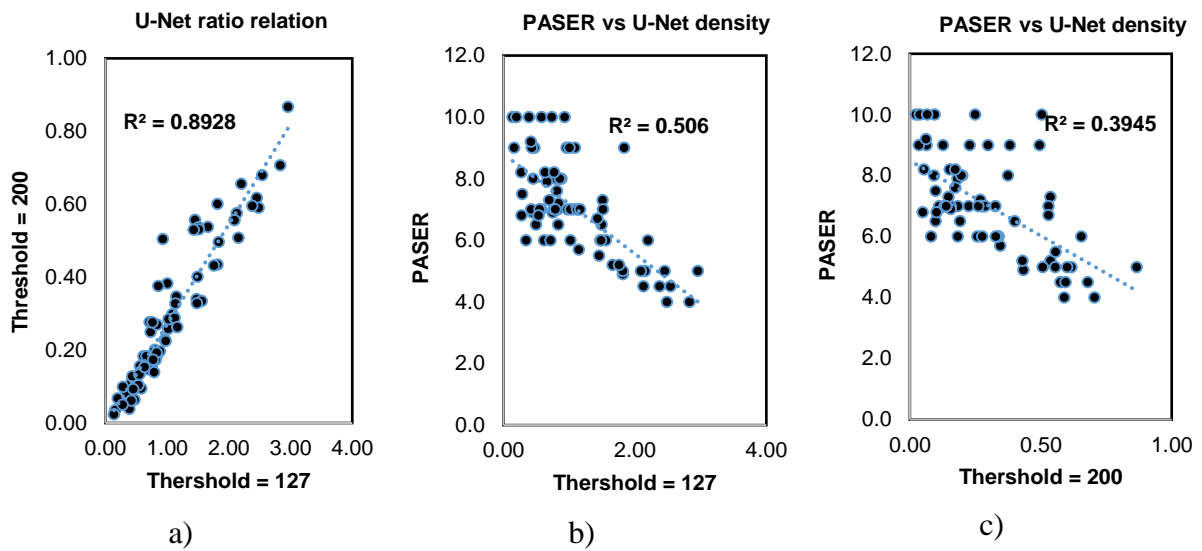

*Figure 17. Correlation of a) distress density with different threshold, b) PASER versus distress density with threshold=127, c) PASER versus distress density with threshold=127*

This model can rank sections based on the density of distresses. However, it cannot differentiate types of distresses and this may lead to a misleading conclusion. For example, this model cannot discriminate sealed cracks from unsealed cracks; hence it could penalize the condition of a section unfairly. Also sealed cracks are usually thicker than the unsealed ones, the ration of white pixels will therefore be higher than images from unsealed crack sections. Hence, the main problem with U-net model ranking is that this model just rely on detecting the density of distresses not types of them. It has been well known that different distresses affect pavement condition differently (**Figure 18**).

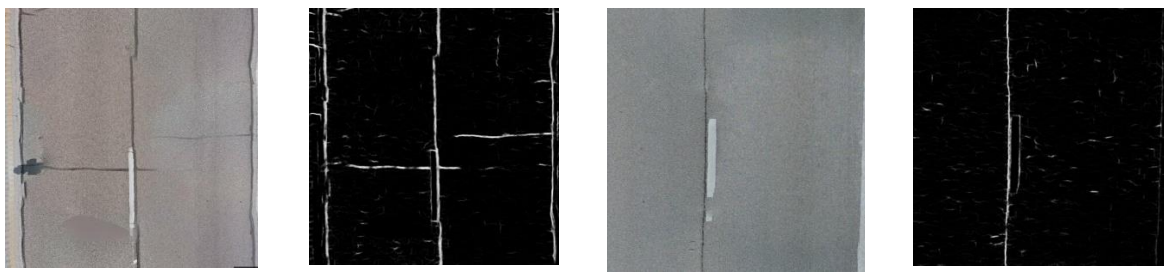

*Figure 18. Example of problems associated with crack density U-Net based model*

According to the limitation of the YOLO and U-net models, we developed a hybrid model including both of these models. Therefore, the type and density of distresses will be considered to develop a new pavement condition index.

*3.4.3. Linear regression PASER prediction model based on hybrid model*

In this part of study, linear regression method was used to develop a prediction model using the explained variables. In order to make the equation similar to PCI approach, the deducted values introduced as the response instead of the measured PASER values. **Table 5** shows the estimated coefficient for the predicted model. Also, P-value were calculated for each variable and the variable number three which is block cracking has the lowest p-value and highest importance in the model. Considering confidence level of 95%, the variable 1, 3, 4, and 8 are significant. **Figure 19** shows the fitted value versus the measured ones for training, testing and validation dataset.

$$PASER\ Predicted = 0.1 * (100 - \sum_{i=1}^{n}(B_i d_i)) \quad [2]$$

where, $d(i)$ = ratio of white pixels to total image size

$d(1)$= Reflective Crack, $d(2)$= Transverse Crack, $d(3)$= Block Crack, $d(4)$= Longitudinal Crack, $d(5)$= Alligator Crack, $d(6)$ = Sealed Reflective Crack, $d(7)$= Lane Longitudinal Crack, $d(8)$= Sealed Longitudinal Crack.

***Table 5.*** *Estimated coefficient for the fitted prediction model using linear regression from R software*

| Variable | Coefficients | Std.Error | t-value | Pr(>\|t\|) | Importance |
|---|---|---|---|---|---|
| d(1) | 100.433 | 13.745 | 7.307 | 2.49E-09 | *** |
| d(2) | -40.603 | 34.906 | -1.163 | 0.2505 | |
| d(3) | 47.709 | 3.312 | 14.405 | 2.00E-16 | *** |
| d(4) | 70.774 | 16.16 | 4.38 | 6.42E-05 | *** |
| d(5) | 40.781 | 27.861 | 1.464 | 0.1498 | |
| d(6) | 14.959 | 11.834 | 1.264 | 0.2123 | |
| d(7) | 20.856 | 16.236 | 1.285 | 0.2051 | |
| d(8) | 7.887 | 3.602 | 2.189 | 0.0335 | * |

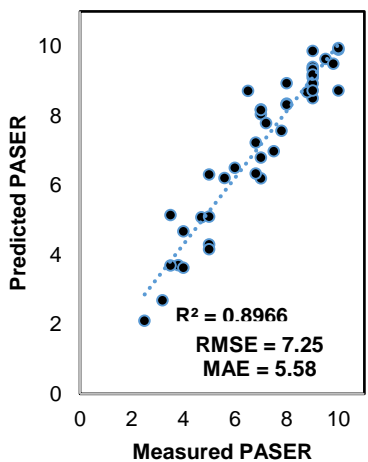
a)

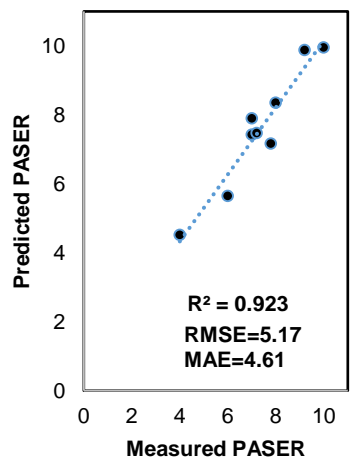
b)

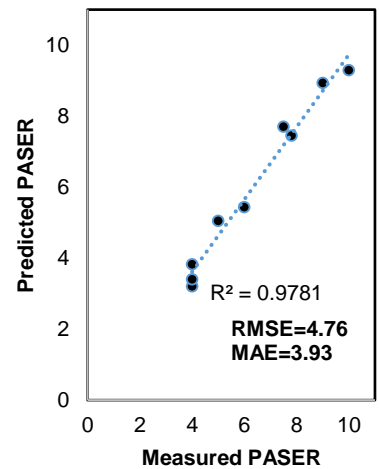
c)

*Figure 19.* Measured PASER versus predicted PASER using linear regression: (a) training data; (b) testing data, and; (c) validation data.

*3.4.4. Weight based PASER prediction model based on hybrid model*

The PCI decision matrix affords particular guidelines for the repairs required based on road classifications. Using PCI, helps to establish a pinpoint for preventive preservation that diminishes road deteriorating before the point that it needs expensive rehabilitation [33].

In this model, different weights were assigned to the distresses, and deduct values were calculated similar to the PCI method (**Table 6** and equation 2). Different weights were considered according to the level of importance of distresses in pavement condition. Afterward, the deduct values were subtracted from 100.

*Table 6.* Distress weights to calculate the predicted PASER using weight method

| Variable | Ratio of white pixels to total image size | Weight |
|---|---|---|
| V1 | Reflective | 0.4 |
| V2 | Transverse | 0.4 |
| V3 | Block | 0.4 |
| V4 | Longitudinal | 0.4 |
| V5 | Alligator | 0.4 |
| V6 | Sealed Reflective | 0.1 |
| V7 | Lane Longitudinal | 0.1 |
| V8 | Sealed Longitudinal | 0.1 |

$$PASER = 0.1 \times (100 - \sum_{i=1}^{n}(w_i d_i)) \qquad [6]$$

where,

d(1)= Reflective Crack, d(2)= Transverse Crack, d(3)= Block Crack, d(4)= Longitudinal Crack, d(5)= Alligator Crack, d(6) = Sealed Reflective Crack, d(7)= Lane Longitudinal Crack, d(8)= Sealed Longitudinal Crack, d(9)= Pothole

**Figure 20** shows the measured PASER values versus the predicted PASER using the PCI weighting method. The coefficient of determination ($R^2$) for the training, testing and validation dataset are 0.87. 0.94, and 0.94. One of the most interesting benefit about the weight-based method over other approaches is that the weights can be adjusted regarding to the case project. For example, for airport pavement monitoring, where reflective cracks are more important than longitudinal cracks. Then desired weights can be assigned to the distresses and a new PASER index will be obtained based on new weights.

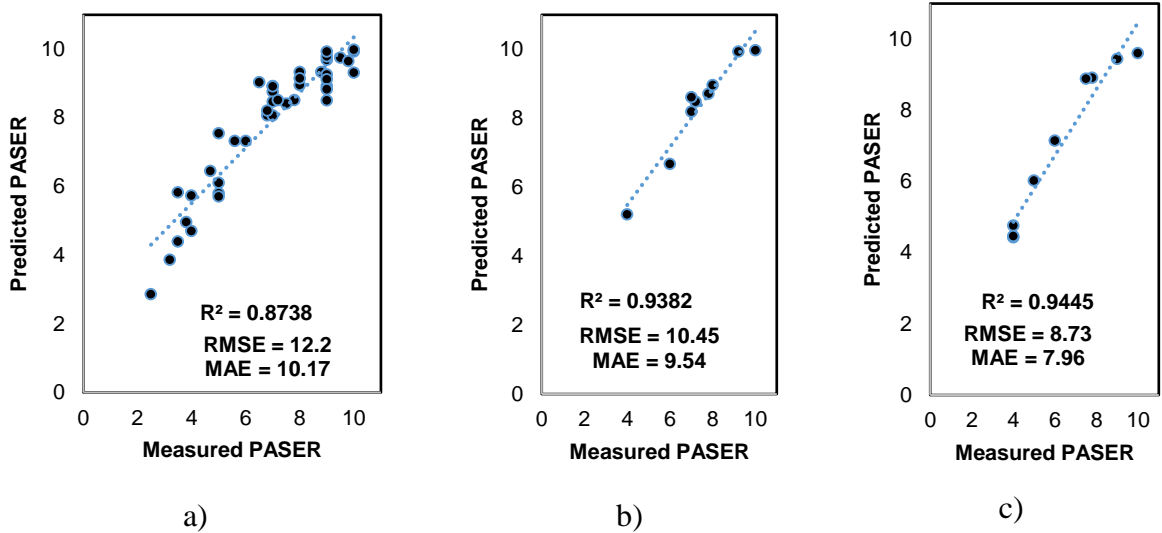

*Figure 20. Measured PASER versus predicted PASER using Weight Method: (a) training, b) testing and (c) Validation dataset.*

## 4. Model Validation

To further evaluate the generalization capability of the PASER prediction models, the developed models were deployed to predict the condition of six pavement sections with different conditions (**Figure 21**). First, a new set of pavement images was extracted for each section from Google maps using our developed software. Second, the images were analyzed with the developed crack detection and segmentation models (YOLO and U-Net) and the ratio of white pixels for each detected distresses to the total image size were calculated. Finally, the PASER values were predicted using weight based and linear regression models for all the extracted images. The images were extracted by 15 m and we considered 5 meter as the view of each image. The moving average of PASER values were calculated by a period of 20 images and plotted in **Figure 22**a and b. **Figure 22**c shows the predicted PASER which developed based on distress classification alone (YOLO model output). As seen in this figure, the fluctuation of the PASER values from YOLO model is less than the ones from hybrid models. This is also expected because the hybrid models are more susceptible and precise due to its dependency on pixels counting. Also **Figure 22**d shows the corresponding measured PASER values extracted from MoDOT's portal. As seen, the predicted PASER values correlate well with the measured values. The rankings are in order according to the measured PASER except swapping MO11 and US24. Contrary to our developed pavement condition prediction models, PASER rating is very straightforward rating without considering any distress quantification in details. The approximate-based nature and human dependency of the PASER procedure, makes it challenging to correlate our machine learning-based pavement condition model with the measured PASER values in some cases. Also, the PASER values from prediction models are fluctuating more than the corresponding measured PASER values due to variety of detected distresses image by image. While measured PASER values are usually constant along 2000 meter which is unrealistic (**Figure 22**).

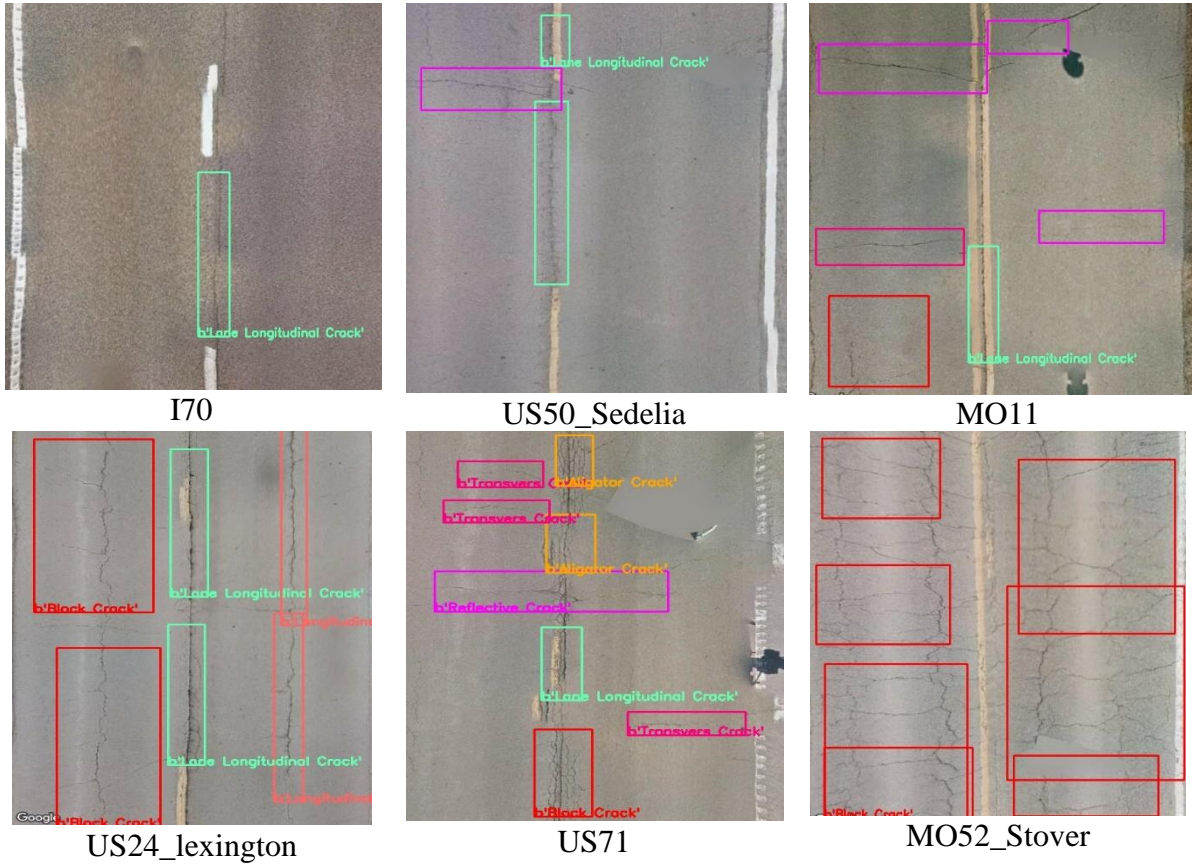

*Figure 21. Pavement condition, a) I70 (good condition), b) US50 , c) MO11, d) US24, e)US71 and f) MO52 (dense-block cracks).*

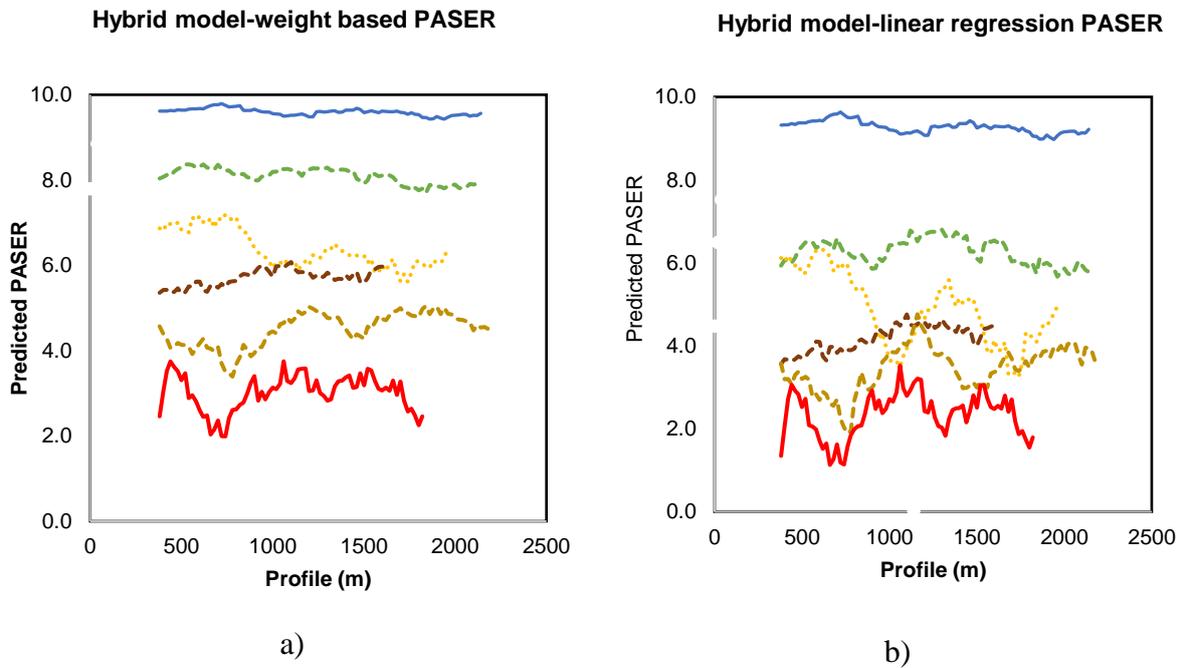

a)

b)

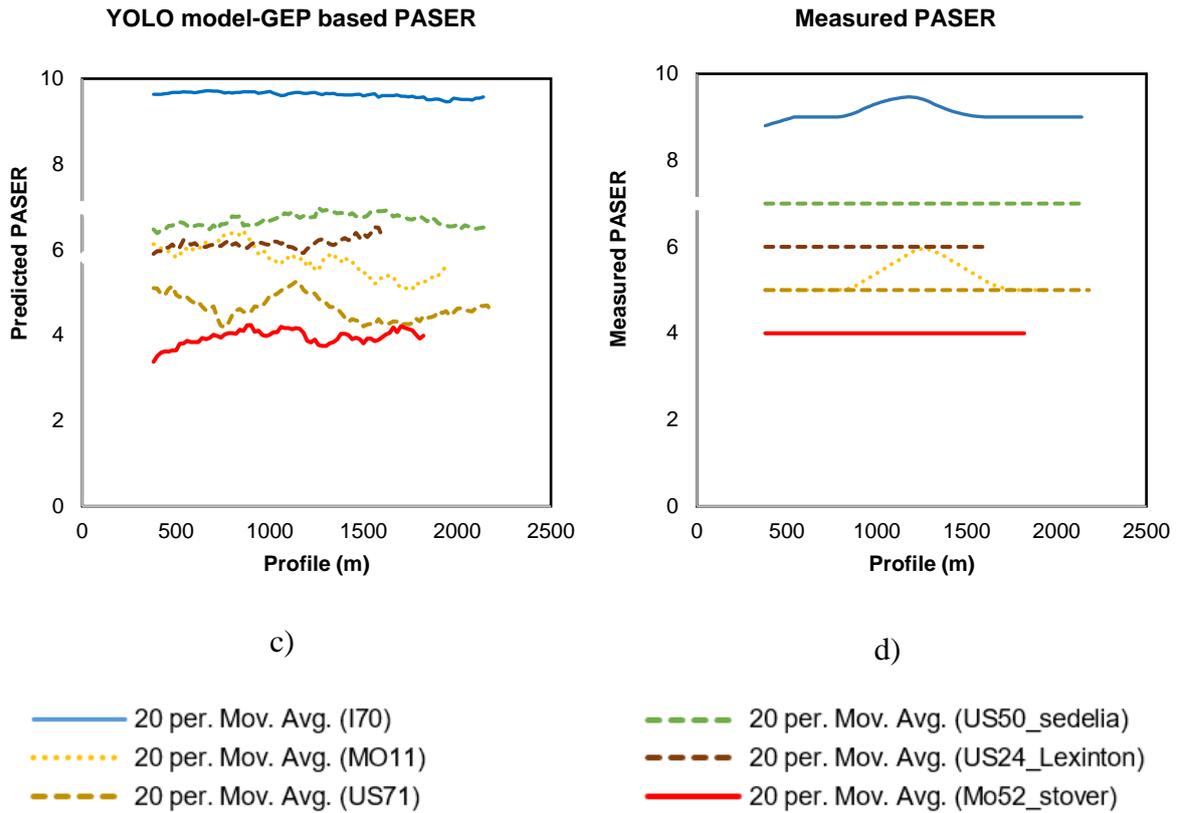

*Figure 22. Changing of Predicted PASER using different models a) Hybrid model-weight based, b) Hybrid model-linear regression, c) YOLO model-GEP based and d) Measured PASER, for six sections alongside pavement profile.*

## 5. Conclusions

In this study, a deep machine learning approach was implemented to predict pavement condition of asphalt-surfaced roadways. Models were trained using a comprehensive road condition dataset consisting of 7,237 images extracted from Google Street-view. The images were annotated with bounding boxes featuring nine different pavement distresses. A state-of-the-art deep learning frameworks were employed to detect, classify and segment nine types of pavement distress automatically. A pre-trained U-Net based model was used to calculate the density of the distresses. The pre-trained model was fine-tuned by using manually annotated road images acquired in a variety of environmental conditions. The model can accurately differentiate non crack-like features including shadows, cars from pavement distresses. In order to develop a model to classify and quantify severity of the distresses, YOLO and U-net model were integrated together as a hybrid model.

Various pavement condition prediction indices were developed based on the detected distresses by YOLO, and hybrid model. Approximately 82 images per section was extracted from 71 different road sections located in the state of Missouri. The distresses detected for all the images using the developed deep-learning crack detection model, and the average number of distresses per image were calculated for all the 71 sections. Afterward, the measured PASER values were extracted from MoDOT portal for all the corresponding road sections. A GEP-based model was developed to predict PASER using the average number of distresses per image as an input. Also, two more PASER prediction models were developed based on the hybrid model outputs. Linear regression, and weight-based prediction models were developed, and the performance of them were verified by high values of coefficients of determination ($R^2$) for all the training, validation, and testing dataset. The results of the sensitivity analysis indicate that block cracking (an extensive distress in the Midway in modern times) is the most effective parameter in explaining the variations in PASER as compared with the other predictor variables. This may be due to the fact that block cracking is a clear indicator of advanced asphalt age and age hardening, both of which are in turn correlated to many other pavement distresses.

For further model validation, six sections with various surface conditions were selected from Missouri roads. The predicted PASER values correlated well with the measured values. However, there was a swap in ranking between two of the cases. Contrary to our proposed pavement condition prediction models, traditional PASER rating is a qualitative-based rating without considering any distress quantity in detail. The approximate and human-based nature of the PASER procedure makes it challenging to correlate our machine learning-based pavement condition model with the measured PASER values in some cases. Also, the predicted PASER values from our models are fluctuating more than the corresponding measured PASER values due to the variety of detected distresses image by image in each section, while PASER values are usually constant over the whole section which is unrealistic.

Finally, the proposed models offer some advantages over traditional pavement monitoring (expensive cost of ARAN vehicles and laser equipment), and as compared to previous deep learning-based models. First, this tool excluded the dependency of PASER to human judgment and made it more accurate. Also this study is the pioneer in concerning about developing a prediction pavement condition index after developing a model to detect the distresses. Second, the models were trained using Google street-view images, which are free and available for virtually all roads in the US and abroad. Third, the models were developed based on comprehensive pavement image dataset which was annotated considering wide variety of common pavement distress types by pavement experts. Finally, the developed models are robust and flexible, cost-effective, and able to predict distress from different camera views towards convenient. The fact that these first-generation models appear to have an acceptable average prediction error suggests that it may be very useful for DOTs and road agencies, as a means to evaluate road sections conditions. This tool could be conveniently employed to evaluate the pavement conditions during its service life and help to make valid decisions for rehabilitation or reconstruction of the roads at the right time.

**Future Studies**

A current limitation in the PASER prediction model reported herein is that it was calibrated using a relatively limited set of PASER results (71 sections). The model performance will be enhanced by updating the model, allowing to continue to learn and ultimately culminate into a well-built and broadly predictive tool, as more data becomes available. It is hoped that collaborations with other research groups and owner-agencies will lead to a vastly larger database, and an even more highly predictive model. It is recommended that a major, national study is justified by the current results, and represents a next logical step forward.

Also, the same idea can be implemented on 3D images and other distresses like rutting can be taken into account in the final pavement condition prediction model. Furthermore, distresses characteristics can be investigated precisely by introducing 3D images.

In order to facilitate the deployment of the proposed model, an integrated software will be developed to incorporate all the steps, including extracting images, analyzing images, predicting pavement condition together.

**Acknowledgments**

The authors would like to thank Missouri Department of Transportation for providing access to MoDOT virtual portal. The findings and conclusions reported herein are those of the authors, and not necessarily those of the MoDOT.

**Conflict of interest**

The authors do not have any conflict of interest with other entities or researchers.